\documentclass{llncs}
\usepackage{makeidx}  % allows for indexgeneration
\usepackage{graphicx}
\usepackage{multirow}
\usepackage{amsmath}
\usepackage{tabularx}
\usepackage{rotating}
\usepackage{color}
\usepackage{adjustbox}
\usepackage{caption}
\usepackage{wrapfig}
\usepackage{soul}

\begin{document}
\frontmatter          % for the preliminaries
%
%\pagestyle{headings}  % switches on printing of running heads
%\addtocmark{Hamiltonian Mechanics} % additional mark in the TOC
%
%
\mainmatter              % start of the contributions
\title{Multiple Instance Learning for Heterogeneous Images: Training a CNN for Histopathology}
%
%\titlerunning{Hamiltonian Mechanics}  % abbreviated title (for running head)
%\author{Anonymous}
%\author{\red{Heather D. Couture\inst{1} \and J.S. Marron\inst{25} \and Charles M. Perou\inst{35} \and \\Melissa A. Troester\inst{45} \and Marc Niethammer\inst{16}}}
\author{Heather D. Couture$^{1}$ \and J.S. Marron$^{2,3}$ \and Charles M. Perou$^{2,4}$ \and \\Melissa A. Troester$^{2,5}$ \and Marc Niethammer$^{1,6}$}
%
%\authorrunning{Ivar Ekeland et al.} % abbreviated author list (for running head)
%Craig Chambers, Kim B. Bruce, and Elisa Bertino}
%
%\institute{*** \\ *** \\ *** \\ *** \\ *** \\ *** }%\\ *** \\ ***}
%\institute{\red{Department of Computer Science, \and Department of Statistics and Operations Research, \and Department of Genetics, \and Department of Epidemiology, \and Lineberger Comprehensive Cancer Center, \and Biomedical Research Imaging Center\\University of North Carolina, Chapel Hill, NC, USA}}
\institute{$^{1}$Department of Computer Science, $^{2}$Lineberger Comprehensive Cancer Center,\\$^{3}$Department of Statistics and Operations Research, $^{4}$Department of Genetics, $^{5}$Department of Epidemiology, $^{6}$Biomedical Research Imaging Center\\University of North Carolina, Chapel Hill, NC, USA}

\maketitle              % typeset the title of the contribution

\begin{abstract}

Multiple instance (MI) learning with a convolutional neural network enables end-to-end training in the presence of weak image-level labels.  We propose a new method for aggregating predictions from smaller regions of the image into an image-level classification by using the quantile function.  The quantile function provides a more complete description of the heterogeneity within each image, improving image-level classification.  We also adapt image augmentation to the MI framework by randomly selecting cropped regions on which to apply MI aggregation during each epoch of training.  This provides a mechanism to study the importance of MI learning.  We validate our method on five different classification tasks for breast tumor histology and provide a visualization method for interpreting local image classifications that could lead to future insights into tumor heterogeneity.

%\keywords{}
\end{abstract}

\section{Introduction}

Deep learning has become the standard solution for classification when a large set of images with detailed annotations is available for training.  When the annotations are weaker, such as with large, heterogeneous images, we turn to multiple instance (MI) learning.  The image (called a bag) is broken into smaller regions (called instances).  We are given a label for each bag, but the instance labels are unknown.  Some form of pooling aggregates instances into a bag-level classification.  By integrating MI learning into a convolutional neural network (CNN), we can learn an instance classifier and aggregate the predictions so the entire system is trained end-to-end \cite{Kraus2016,Sun2016,Jia2017}.

\begin{figure}
\centering
\includegraphics[width=4.8in]{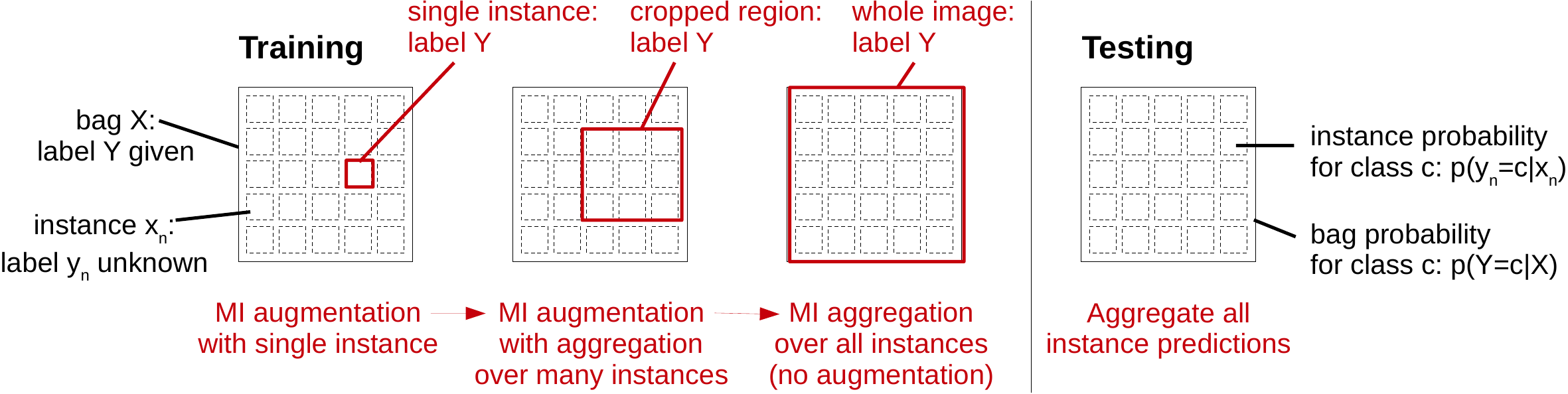}
\vspace{-20pt}
\caption{In MI learning, each bag contains one or more instances.  Labels are given for the bag, but not the instances.  MI augmentation is a technique to provide additional training samples by randomly selecting a cropped image region and the instances within it.  When the bag label is applied to a small number of instances, it is weak because this small region may not be representative of the bag class.  Applying the bag label to larger cropped regions provides a stronger label, while still providing benefit from image augmentation.  Training with the whole image maximizes the opportunity for MI learning, but restricts the benefits of image augmentation.  {\it At test time, the whole image is processed and the predictions from all instances are aggregated into a bag prediction.}}
\vspace{-12pt}
\label{fig_miaug}
\end{figure}

We propose a more general approach for aggregating instance predictions that looks at the full distribution by pooling with the quantile function (QF) and learning how much heterogeneity to expect for each class.  As data augmentation is especially critical in training large CNNs, we also created an augmentation technique for training MI methods with a CNN (Fig. \ref{fig_miaug}).  Through MI augmentation, we study the importance of the MI formulation during training.

Using MI learning to make class predictions over smaller regions of the image provides insight into how different parts of the image contribute to the classification.  Visualizing the instance predictions provides a method of interpretability that we demonstrate on a data set of breast tumor tissue microarray (TMA) images stained with hematoxylin and eosin (H\&E) by predicting grade, receptor status, and subtype.  Some of these tasks are not previously known to be achievable from H\&E alone.  Our quantitative results conclude that the MI component is critical to successful classification, demonstrating the importance of accounting for heterogeneity.  This method could provide future insights into tumor heterogeneity and its connection with cancer progression \cite{Hiley2014,McGranahan2015}.

{\bf Contributions.} 1) A more general MI aggregation method that uses the quantile function for pooling and learns how to aggregate instance predictions.  2) An MI augmentation technique for training MI methods.  3) Exploration of single instance and MI learning on a continuous spectrum, demonstrating the importance of MI learning on heterogeneous images.  4) Evaluation on a large data set of 1713 patient samples (5970 images), showing significant gains in classifying breast cancer TMAs.  5) A method for visualizing the predictions of each instance, providing interpretability to the method.

\section{Background}

\subsubsection{Aggregating Instance Predictions.} A {\it permutation invariant} pooling of instances is needed to accommodate images of different sizes, whereas a fully connected neural network cannot.  Existing pooling approaches are very aggressive; they compute a single number rather than looking at the distribution of instance predictions.  Most MI applications use the maximum, which works well for problems such as cancer diagnosis where, if there is a small amount of tumor, the sample is labeled as cancerous \cite{Kandemir2014_CMIG,Xu2014_MIA}.  A smooth approximation, such as the generalized mean or noisy-OR, provide better convergence in a CNN \cite{Kraus2016,Sun2016,Jia2017}.  For other tasks, a majority vote, median, or mean is more appropriate.  We include more of the distribution by pooling with the QF and learning a mapping to the bag class prediction, improving the classification accuracy.  Our proposed method of quantile aggregation learns how to predict the bag class from instance predictions and so could provide a solution when the most suitable aggregator is unknown.  The QF is a new general type of feature pooling that could provide an alternative to max pooling in a CNN.

\subsubsection{Training MI Methods with a CNN.} Image augmentation is commonly applied in training a CNN by randomly cropping large portions of each image during each epoch.  At test time, the whole image is used.  We propose MI augmentation, in which a subset of instances is randomly selected from each bag during each epoch.  Instances are always the same size, but we choose how many instances to aggregate over.  In selecting the number of instances, there are two extremes: a single instance vs. the whole bag.  In the former, the bag label is assigned to each instance and is often called single instance learning.  In the latter, MI aggregation is incorporated while training the bag classifier as in other MI methods \cite{Andrews2002,Hou2016}.  Comparison studies have found little or no improvement from these MI methods on some data sets \cite{Vanwinckelen2016,Wang2018}.  We found MI learning to be very beneficial and show that it is critical in dealing with heterogeneous data.

\section{Multiple Instance Learning with a CNN}

\begin{figure}
\centering
\includegraphics[width=4.4in]{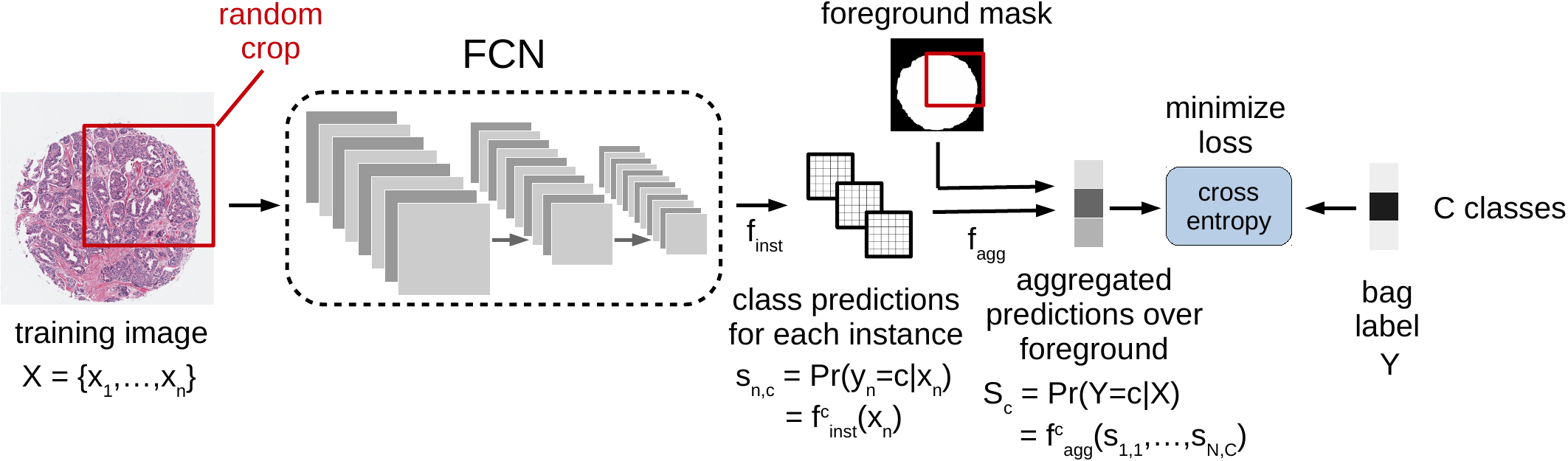}
\vspace{-10pt}
\caption{During training, a cropped region of a given size is randomly selected.  An FCN is applied to predict the class, producing a grid of instance predictions.  The instance predictions are aggregated over the foreground of the image (as indicated by the foreground mask) using quantile aggregation to predict the class of the cropped image region.  With a cross entropy loss applied, backpropagation then learns the FCN and aggregation function weights.  At test time, the whole image is used.}
\vspace{-12pt}
\label{fig_mi_aug_cnn}
\end{figure}

We denote a bag by $X$, its label by $Y \epsilon\{1,2,...,C\}$, and the instances it contains by $x_{n}$ for $n=1,...,N$.  The instance labels $y_n$ are unknown.  On a novel sample, an instance classifier $f^c_{inst}$ predicts the probability of each class $c$ and a function $f^c_{agg}$ aggregates these instance probabilities into a bag probability:
\vspace{-5pt}
\[ s_{n,c} = f^c_{inst}(x_n) = \Pr(y_{n}=c|x_n) ~~~~~~ S_c = f^c_{agg}(s_{1,1},...,s_{N,C}) = \Pr( Y = c| X) .\vspace{-5pt}\]
MI learning can be implemented with many different types of classifiers \cite{Andrews2002,Kandemir2014_CMIG,Vanwinckelen2016}.  When implemented as a CNN, a fully convolutional network (FCN) forms the instance classifier $f_{inst}$, followed by a global MI layer for instance aggregation $f_{agg}$.  The FCN consists of convolutional and pooling layers that downsize the representation, followed by a softmax operation to predict the probability for each class.  For an input image of size $w \times w \times 3$, the FCN output is $w_d \times w_d \times C$.  An instance is defined as the receptive field from the original image used in creating a point in this $w_d \times w_d$ grid; the instances are overlapping.  The MI aggregation layer takes the instance probabilities and the foreground mask for the input image (downscaled to $w_d \times w_d$), thereby aggregating over only the foreground instances.  Figure \ref{fig_mi_aug_cnn} provides an overview.

\section{Multiple Instance Aggregation}
\label{sec_miagg}

Instance predictions can be used to form a bag prediction in different ways.  The bag prediction function should be invariant to the number and spatial arrangement of instances, so some pooling of predictions is needed. 
Mean aggregation is well suited for global pooling as it is permutation invariant and can incorporate a foreground mask for the input image.  Denoting the mask as $M$ and its value for each instance as $m_n \epsilon\{0,1\}$, the mean aggregation function is
$S_c = f^c_{\text{mean agg}}(s_{1,1},...,s_{N,C}) = \frac{\sum_{n=1}^N m_n s_{n,c} }{\sum_{n=1}^N m_n}$ .

Mean pooling incorporates predictions from all instances, but a lot of information is lost in compressing to a single number.  A histogram is a more complete description of the probability distribution, but is dependent upon a suitable bin width.  Alternatively, the QF (inverse cumulative distribution) represents the boundary points between fractions of the population, providing a better discretization \cite{Broadhurst2008}.  We propose quantile aggregation to provide a more complete description of the instance predictions in a bag.  If the instance predictions for class $c$ are represented by $S_c = \{s_{1,c},...,s_{N,c}\}$, then the $q$-th $Q$-quantile is the value $z$ such that $\Pr(S_c \leq z) = (q-0.5)/Q$.  To pool with the QF, we first sort $S_c$ and exclude instances not in the foreground, leaving the set $\tilde{S}_c=\{\tilde{s}_{1,c},...,\tilde{s}_{\tilde{N},c}\}$.  The sorted values in $\tilde{S}_c$ are used to extract the QF vector for each class $c$ as $z_c = [z_{1,c},...,z_{Q,c}]$ where $z_{q,c} = \tilde{s}_{\lceil \tilde{N}(q-0.5)/Q) \rceil,c}$.
The QF vectors for all classes are concatenated as $Z=[z_1,...,z_C]$.  We then use a softmax function operating on $Z$ to predict the bag class.  The QF from all classes is used in order to learn the interaction of different subtypes in a bag.  Backpropagtion through the QF operates in a similar manner to max pooling by passing the gradient back to the instance that achieved each quantile.

\section{Training with Multiple Instance Augmentation}

Image augmentation by random cropping is an important technique for creating extra training samples that helps to reduce over-fitting.
We propose an augmentation strategy for MI methods to increase the number of training samples by randomly selecting a different subset of instances for each epoch.  We randomly crop the image to select the set of instances, such that each crop contains at least 75\% foreground according to the foreground mask.  It is important to note that the image is never resized and the instance size remains constant.  For each crop size chosen, the FCN is applied to the cropped image at full resolution.  MI augmentation is a strategy used during training.  {\it As the MI aggregation layer is invariant to input size, the entire image and all its instances are always used at test time.}

\section{Experiments}

\subsubsection{Data Set.}

Our data set consists of 1713 patient samples from the Carolina Breast Cancer Study, Phase 3 \cite{Troester2018}.  There are typically four 1.0 mm cores per patient in the TMA, with a total of 5970 cores.  Each core is selected from the H\&E-stained whole slide by a pathologist such that it contains a substantial amount of tumor tissue.  Each image has a diameter of around 2400 pixels and a maximum of 3500 pixels.  One sample core is shown in Fig. \ref{fig_mi_aug_cnn}.  We use a random subset of half the patients for training and the other half for testing.  Classification accuracy is measured for five different tasks, some of them multi-class: 1) histologic subtype (ductal or lobular), 2) estrogen receptor (ER) status (positive or negative), 3) grade (1, 2, or 3), 4) risk of recurrence score (ROR) (low, intermediate, or high), 5) genetic subtype (basal, luminal A, luminal B, HER2, or normal-like).  Ground truth for histologic subtype and grade are from a pathologist looking at the original whole slide.  ER status is determined from immunohistochemistry, genetic subtype from the PAM50 array \cite{Parker2009}, and ROR from the ROR-PT score-based method \cite{Parker2009}.

\subsubsection{Implementation Details.}

The TMA images are intensity normalized to standardize the appearance across slides \cite{Niethammer2010}.
The hematoxylin, eosin, and residual channels are extracted from the normalization process and used as the three-channel input for the rest of our algorithm.  A binary mask distinguishing tissue from background is also provided as input.

We use the pre-trained CNN AlexNet \cite{Krizhevsky2012} and fine-tune with the MI architecture shown in Fig. \ref{fig_mi_aug_cnn}.  All five tasks are equally weighted in a multi-task CNN as shared features help to reduce over-fitting.  For each patient, ground truth labels are available for most tasks.  The cross entropy loss is adjusted to ignore patients missing a label for a particular task.

In addition to MI augmentation, we randomly mirror and rotate each training image.  To accommodate the larger cropped image sizes in GPU memory, we reduce the batch size.  A typical image with tissue of diameter 2400 pixels produces a $68 \times 68$ grid of instances.  After applying the foreground mask, there are roughly 3600 instances.  $Q=15$ quantiles are used in all experiments.  There are typically four core images per patient; we assign the patient label to each during training and, at test time, take the mean prediction across the images.  Further MI learning could be done to address the multiple core images per patient, however our current focus is only on MI learning within each image.

\subsubsection{MI Augmentation and the Importance of MI Learning.}

We study the effect of MI learning on large images by selecting the cropped image size for training.  The smallest possible size is $227 \times 227$ (the input size for AlexNet), consisting of a single instance.  When the bag label is applied to each instance during training, this is called single instance learning.  Alternatively, a larger cropped region of size $w \times w$ can be selected; we test multiples of 500 up to 3500 and use mean aggregation in this experiment.  By assigning the bag label to this larger cropped region during training and keeping the instance size constant, we perform MI learning.  Multiple random crops are obtained from each training image such that roughly the same number of pixels is sampled for each crop size (i.e., the whole image for the largest crop size of 3500, $\frac{3500^2}{w^2}$ random crops for a training crop of size $w$).  For the largest crop size, the whole image is used without MI augmentation.  Random mirroring and rotations are used for augmentation at all crop sizes.  At test time, the whole image is always used, with the bag prediction formed by aggregating across all instances.

\begin{wrapfigure}[14]{r}{0.5\textwidth}
  %\vspace{-20pt}
  \vspace{-25pt}
  \includegraphics[width=2.4in]{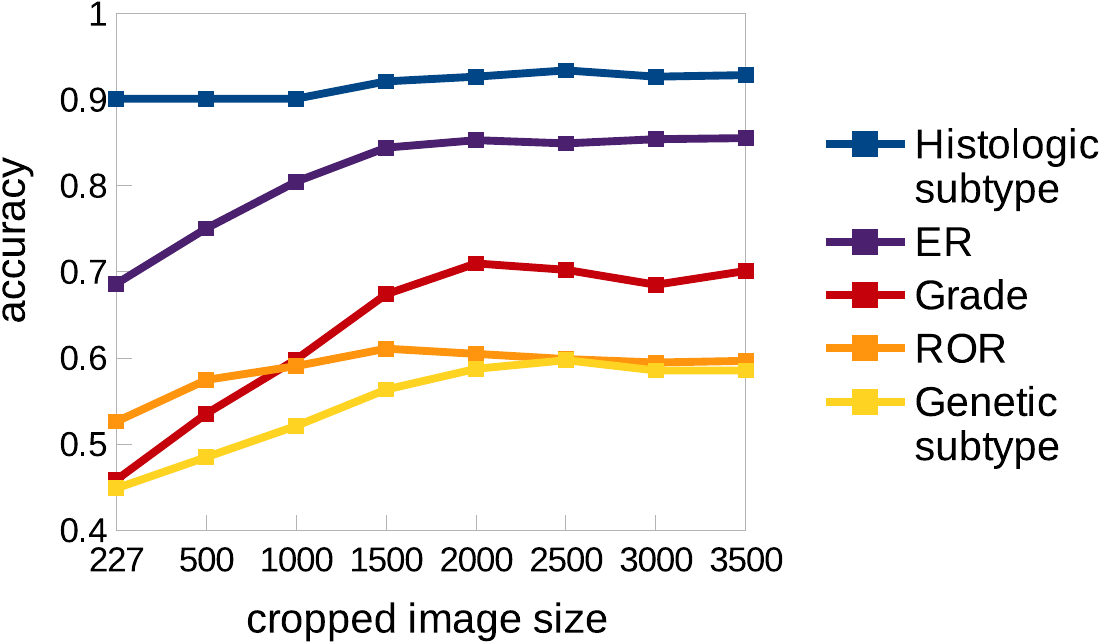}
  \vspace{-18pt}
  \caption{Classification accuracy using mean aggregation as the number of instances (cropped image size) used for training is increased, while keeping instance size constant.}
  \label{fig_plots}
\end{wrapfigure}

Figure \ref{fig_plots} shows that larger crop sizes {\it for training} significantly increases classification accuracy ($p<10^{-3}$ with McNemar's test for w=500 vs. w=1500 on all tasks).  The benefits level off for larger crops.
As GPU memory requirements increase for larger crop sizes, selecting an intermediate crop size provides most of the benefits of MI augmentation.

Although it should not be surprising that a larger crop size at training works better, the magnitude of improvement is very significant.  If the images were homogeneous (at the scale of a single instance, $w=227$), then applying the bag label to each instance should produce a classification accuracy similar to when MI aggregation over the whole image is used during training.  This is clearly not the case in Fig. \ref{fig_plots}.  For example, ER status increases from 68.6\% to 85.6\% when applying MI learning over the whole image.  {\it This demonstrates the importance of MI learning and the effect of heterogeneity.}  Our data set consists of cores selected from a whole slide by a pathologist.  MI learning may be even more crucial when classifying larger and more heterogeneous images like whole slides.

\subsubsection{MI Aggregation.}

\begin{wraptable}{r}{0.55\textwidth}
  %\vspace{-67pt}
  \vspace{-54pt}
  \adjustbox{width=2.7in}{
  \begin{tabular}{lcccc}
    \hline
    Task & Max & Mean & Quantile \\
    \hline
    Histologic subtype & .898 (.004) & .931 (.004) & .952 (.003) \\
    ER & .683 (.006) & .833 (.008) & .841 (.006) \\
    Grade & .408 (.019) & .680 (.003) & .676 (.006)  \\
    ROR & .542 (.010) & .595 (.003) & .582 (.008) \\
    Genetic subtype & .321 (.032) & .548 (.006) & .544 (.003) \\
    \hline
  \end{tabular}
  }
  \vspace{-8pt}
  \caption{Average classification accuracy for different types of MI aggregation.  The standard error is in brackets.}
  \vspace{-25pt}
  \label{tbl_agg}
\end{wraptable}
We compared aggregation methods by training our model on a crop size $w=2000$ and taking the average classification accuracy over four runs.  Table \ref{tbl_agg} shows that mean and quantile aggregation both significantly outperform max ($p<10^{-8}$ with McNemar's test).  
While quantile aggregation performance is similar to mean for some tasks, a significant increase in performance (93.1\% to 95.2\%) is observed for predicting the histologic subtype of ductal vs. lobular ($p<10^{-10}$ with McNemar's test).  This improvement is due to quantile aggregation predicting the bag class from a more complete view of the instance predictions using QF pooling, thereby capturing the heterogeneity.

\subsubsection{Heterogeneity.}

\begin{figure}
  \vspace{-10pt}
  \centering
\begingroup
  \setlength{\tabcolsep}{1.1pt}
  \renewcommand{\arraystretch}{0.5}
\begin{tabular}{ccccccccccc}
  Histology &
  Histologic &&
  \multirow{2}{*}{ER status} &&
  \multirow{2}{*}{Grade} &&
  \multirow{2}{*}{ROR} &&
  Genetic\\
  image & subtype & & & & & & & & subtype \\
  \includegraphics[height=0.45in]{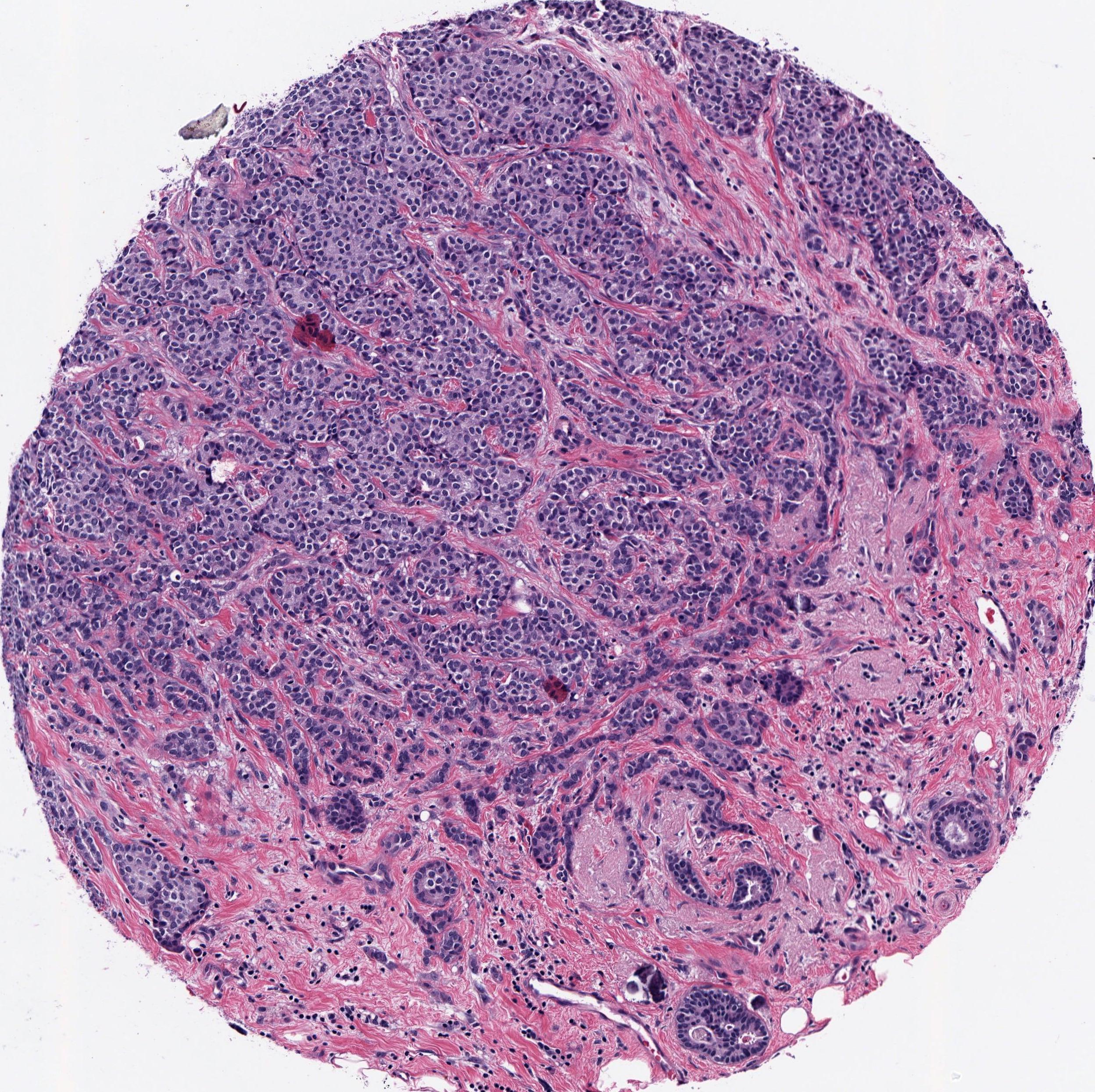} &
  \includegraphics[height=0.45in]{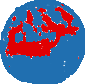} &
  \includegraphics[scale=0.5]{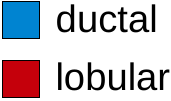} &
  \includegraphics[height=0.45in]{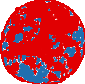} &
  \includegraphics[scale=0.5]{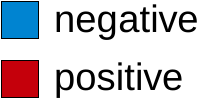} &
  \includegraphics[height=0.45in]{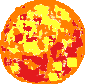} &
  \includegraphics[scale=0.5]{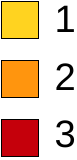} &
  \includegraphics[height=0.45in]{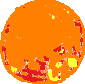} &
  \includegraphics[scale=0.5]{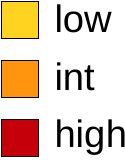} &
  \includegraphics[height=0.45in]{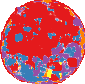} &
  \includegraphics[scale=0.5]{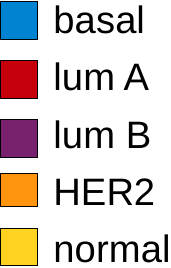} \\
\end{tabular}
\endgroup
\vspace{-10pt}
\caption{Visualization of instance predictions for a sample with ground truth labels of ductal, ER positive, grade 1, low ROR, and luminal A.}
  \vspace{-20pt}
\label{fig_vis}
\end{figure}

\begin{figure}
  \vspace{-10pt}
  \centering
  \setlength{\tabcolsep}{10pt}
  \begin{tabular}{cc}
    Grade 1 vs. 3 & Genetic Subtype Basal vs. LumA \\
\includegraphics[height=1.5in]{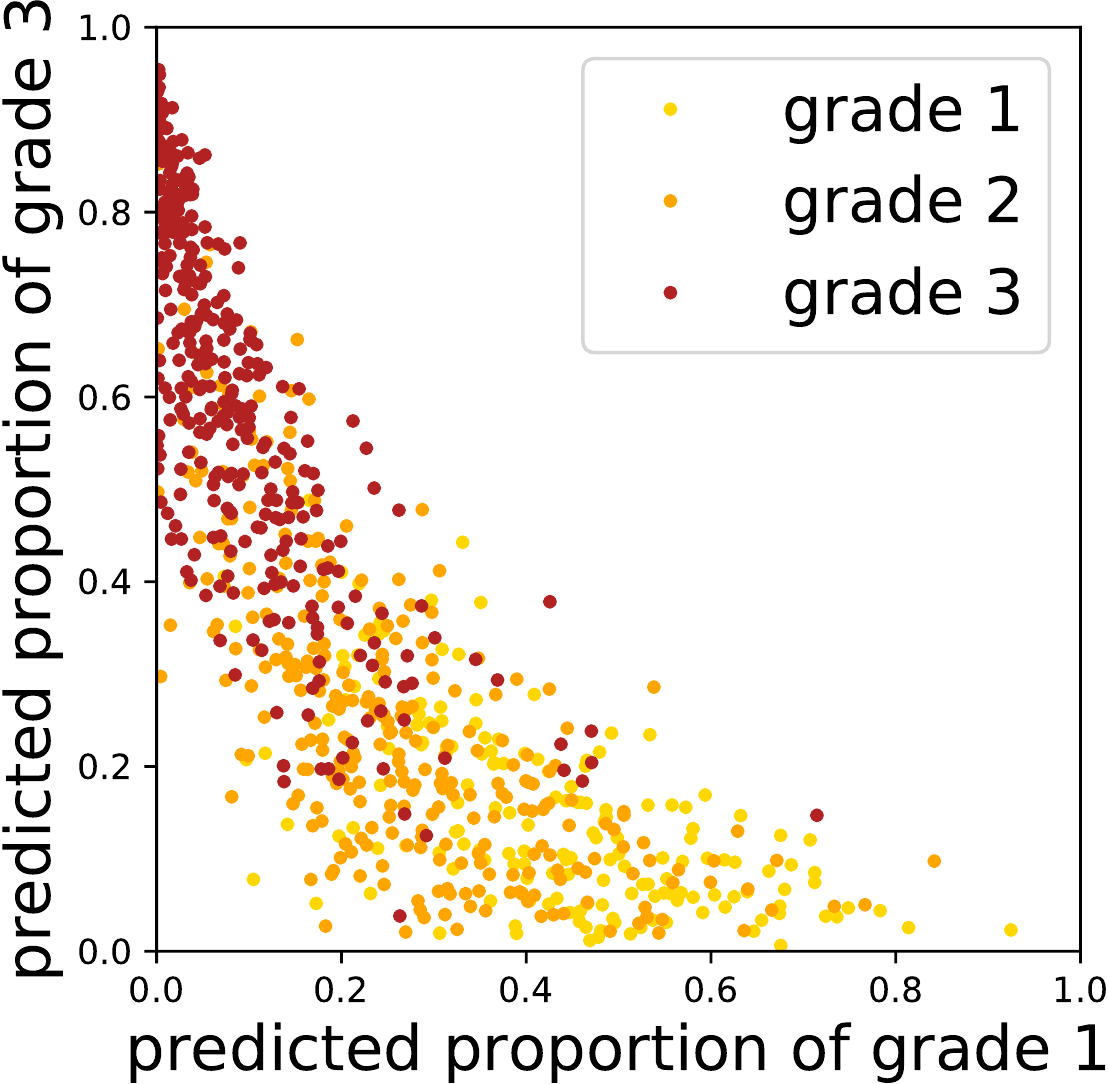} &
\includegraphics[height=1.5in]{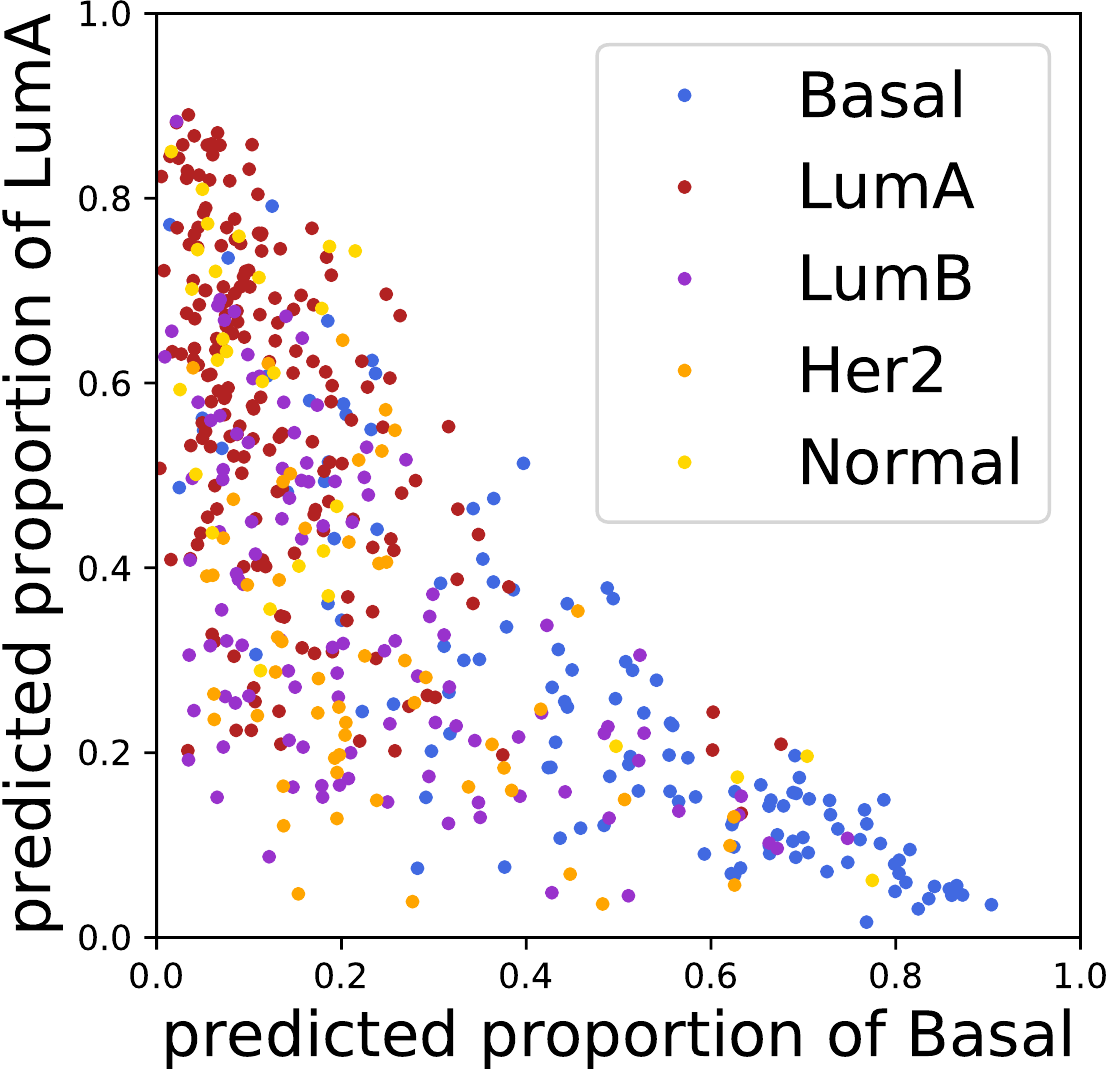} \\
\end{tabular}
\vspace{-5pt}
\caption{Predicted heterogeneity for grade 1 vs. 3 and genetic subtype basal vs luminal A.  The predicted proportion for each class is calculated as the proportion of instances in the sample predicted to be from each class.  Test samples for all classes are plotted.}
  \vspace{-25pt}
\label{fig_het_plots}
\end{figure}

By computing the class predictions for each instance, we get an idea of each region's contribution to the classification.  Figure \ref{fig_vis} provides a visualization for a sample image where the instance predictions are colored for each class.  The $w=2000$ crop size was used for this example.  With the same computation performed over the whole test set, we calculated the proportion of instances predicted to belong to each class.  Figure \ref{fig_het_plots} plots the results for grade 1 vs. 3 and genetic subtype basal vs. luminal A.  Heterogeneity is expected for grade, as the three tumor grades are not discrete, but a continuous spectrum from low to high.  On the other hand, the level of heterogeneity to expect for genetic subtype is unknown because no studies have yet assessed genetic subtype from multiple samples within the same tumor.  The graph shows a continuous spectrum from basal to luminal A.  The luminal B, HER2, and normal samples lie mostly on the luminal A side, but with some mixing into the basal side.

\section{Discussion}

We have shown that MI learning while training a CNN is critical in achieving high classification accuracy on large, heterogeneous images.  Even with a small number of labeled samples, our model was successful in fine-tuning the AlexNet CNN because of the large size of the images providing plenty of opportunity for MI augmentation.  The impact of MI learning indicates that accommodating image heterogeneity is essential.  While aggregating instance predictions with the mean is sufficient for some tasks, quantile aggregation produces a significant improvement for others.  Instance-level predictions will enable future work studying tumor heterogeneity, perhaps leading to biological insights of tumor progression.

%\vspace{-5pt}
\paragraph{\bf\footnotesize Acknowledgments.}  \begin{sloppypar}\footnotesize{ This work was supported by a grant from the UNC Lineberger Comprehensive Cancer Center funded by the University Cancer Research Fund (LCCC2017T204), NCI Breast SPORE program (P50-CA58223), and the Breast Cancer Research Foundation.}\end{sloppypar}
%Removed for anonymization.

\bibliographystyle{splncs03}
\bibliography{micnn}

\end{document}